\documentclass{article}
\usepackage{iclr2021_conference_main,times}

\usepackage{microtype}
\usepackage{appendix}
\usepackage{graphicx}
\usepackage{caption}
\usepackage{subcaption}
\usepackage{booktabs}
\usepackage{multirow}
\usepackage{varwidth}
\usepackage{soul}
\usepackage{clrscode3e}
\usepackage{makecell}
\usepackage{paralist}
\usepackage{natbib}
\usepackage{bm} 
\usepackage{verbatim}
\usepackage[export]{adjustbox}

\usepackage{tikz}
\usepackage{amsmath, amssymb,mathtools}

\usetikzlibrary{positioning,decorations.pathmorphing,calc,shapes}

\usepackage{booktabs} 

\definecolor{mygreen}{HTML}{006400}
\definecolor{mynavy}{HTML}{000080}
\definecolor{vividred}{HTML}{E60B42}

\newcommand{\hs}[1]{\textcolor{red}{\textbf{hs}: #1}}

\usepackage[unicode=true]{hyperref}
\usepackage{cleveref}

\usepackage{url}

\title{Persistent Message Passing}

\author{Heiko Strathmann, Mohammadamin Barekatain, Charles Blundell and Petar Veli\v{c}kovi\'{c} \\
DeepMind \\
\texttt{\{strathmann,barekatain,cblundell,petarv\}@google.com}
}

\iclrfinalcopy 
\begin{document}

\maketitle
\vspace{-.5cm}
\begin{abstract}
Graph neural networks (GNNs) are a powerful inductive bias for modelling algorithmic reasoning procedures and data structures.
Their prowess was mainly demonstrated on tasks featuring Markovian dynamics, where querying any associated data structure depends only on its \emph{latest} state.
For many tasks of interest, however, it may be highly beneficial to support efficient data structure queries dependent on previous states. This requires tracking the data structure's evolution through time, placing significant pressure on the GNN's latent representations.
We introduce Persistent Message Passing (PMP), a mechanism which endows GNNs with capability of querying past state by explicitly \emph{persisting} it: rather than overwriting node representations, it creates new nodes whenever required.
PMP generalises out-of-distribution to more than $2\times$ larger test inputs on dynamic temporal range queries, significantly outperforming GNNs which overwrite states.
\end{abstract}

\section{Introduction}
Graph neural networks (GNNs) are one of the the most impactful approaches to processing data over irregular domains (e.g., quantum chemistry \citep{klicpera2020directional}, drug discovery \citep{stokes2020deep}, social network analysis \citep{pal2020pinnersage} and physics simulations \citep{pfaff2020learning}). \emph{Combinatorial optimisation} tasks \citep{nair2020solving} have been used to test the limits of GNNs particularly when \emph{extrapolation} is required. Extrapolation, in general, is now known to occur in GNNs only under stringent conditions on their architecture and input featurisation \citep{xu2020neural}. 

\emph{Algorithmic reasoning} \citep[Section 3.3.]{cappart2021combinatorial} has been the source of much of the knowledge of how to build extrapolating GNNs. This emerging area of research seeks to emulate each iteration of classical algorithms \citep{cormen2009introduction} directly within neural networks. Through the lens of algorithmic alignment \citep{xu2019can}, GNNs can be constructed that closely mimic \emph{iterative computation} \citep{velivckovic2019neural,tang2020towards}, \emph{linearithmic sequence processing} \citep{freivalds2019neural}, and \emph{pointer-based data structures} \citep{velivckovic2020pointer}. Also such approaches are capable of \emph{strongly} generalising \citep{yan2020neural} and \emph{data-efficient} planning \citep{deac2020xlvin}.

We study GNN-based reasoners acting on inputs that \emph{dynamically change}  over time. This setup was previously studied briefly \citep{velivckovic2020pointer} and only featured \emph{Markovian} querying: only using the \emph{latest} version of the data. 
In the non-Markovian case, queries require knowledge of \emph{previous} versions of the data. For example, during search, a particular path of a search tree may be expanded but later backtracked. Efficient queries over histories of states serve as a basis for agents solving POMDPs. In economics, historical data are revised and these revisions themselves are of interest.

Most GNNs are designed with Markovian querying in mind: latent representations are \emph{overwritten} in every step, and the last representation is used to answer queries. 
This overloads GNNs' latents, as \emph{all} past snapshots of the data must be represented within them. Further, \citet{hinton2021how} recently makes a strong case for \emph{replicated} embeddings. We propose \textbf{Persistent Message Passing} (PMP), replacing overwriting with \emph{persisting}: when updating a node's state, a \emph{copy} of that node is preserved for later use. To avoid excessive memory usage, PMP has a mechanism to select which nodes to persist.

PMP effectively provides GNNs with an \emph{episodic memory} \citep{pritzel2017neural} of their previous computations, just as in social networks where dynamically changing graph data may necessitate explicit memory modules \citep{rossi2020temporal}. PMP aligns with the broad class of \emph{persistent data structures} \citep{driscoll1989making}, which further \emph{expands} the space of general-purpose algorithms that can be neurally executed. We show on \emph{dynamic range querying} that our method provides significant benefits to overwriting-based GNNs, both in- and out-of-distribution.

\begin{figure}
\vspace{-1em}
     \centering
     \begin{subfigure}[b,valign=t]{0.24\textwidth}
         \centering
         \includegraphics[width=\linewidth]{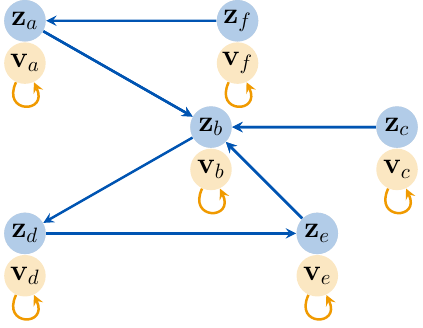}
         \caption{}
         \label{fig:y equals x}
     \end{subfigure}
     \hfill
     \begin{subfigure}[b,valign=t]{0.24\textwidth}
         \centering
         \includegraphics[width=\linewidth]{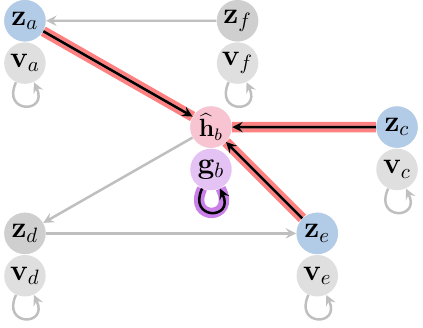}
         \caption{}
         \label{fig:three sin x}
     \end{subfigure}
     \hfill
     \begin{subfigure}[b,valign=t]{0.24\textwidth}
         \centering
         \includegraphics[width=\linewidth]{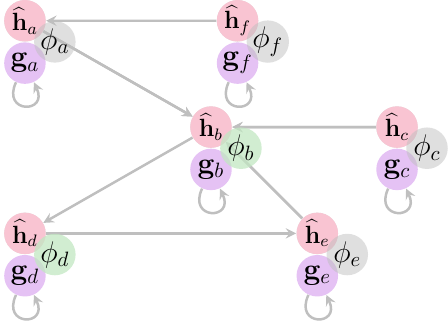}
         \caption{}
         \label{fig:five over x}
     \end{subfigure}
     \hfill
    \begin{subfigure}[b,valign=t]{0.24\textwidth}
         \centering
         \includegraphics[width=\linewidth]{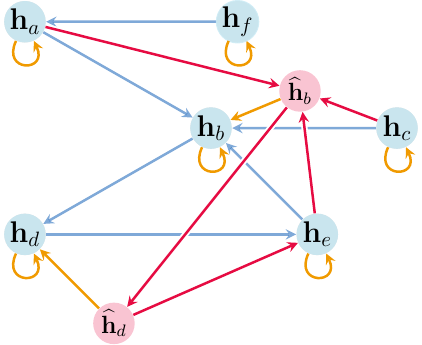}
         \caption{}
         \label{fig:five over x}
     \end{subfigure}
        \caption{\textbf{Overview of PMP}. \textbf{(a)} Derived representations for operation features $\bm{z}$ (blue) and relevance features $\bm{v}$ (yellow). Initially, relevance nodes point to themselves and the operation connectivity is provided. \textbf{(b)} Node $b$ receives messages from its neighbours, leading to next-step candidate latents $\widehat{\bm{h}}_b$, and relevance latents $\bm{g}_b$. \textbf{(c)} Persistency mask prediction selects nodes $b$ and $d$. \textbf{(d)} Selected nodes are copied (red) while original nodes (blue) remain untouched. Connectivity of new nodes mimics that of their predecessors, and relevance connectivity (yellow) points to persisted ``old'' nodes.}
       \label{fig:overview}
       \vspace{-1em}
\end{figure}

\section{Problem setup and PMP architecture}
\label{sec:pgn}

Consider a sequential supervised learning setting:
We are given a set of $n$ entities, along with connectivity information between them.
At each time step, we perform \emph{operations}, which both may change the entities' states and emit an output that we aim to predict.
Each operation may further be associated with a specific \emph{snapshot} of the entities, allowing operations over past states.

More formally, we are given an adjacency matrix, 
$\bm{\Pi}\in\{0,1\}^{n\times n}$, and a sequence of input pairs $(\mathcal{E}^{(1)}, s^{(1)}),(\mathcal{E}^{(2)},  s^{(2)}),\ldots$ where $\mathcal E^{(t)}=(\bm{e}_1^{(t)},  \dots, \bm{e}_n^{(t)})$ are  feature vectors (operation), and $s^{(t)}\in\{1,\dots,t-1\}$ are integers (snapshot indices).
The task is to predict operation outputs $\bm{y}^{(t)}$ given $(\mathcal{E}^{(t)}, s^{(t)})$ where we have \emph{persistency}: $(\mathcal{E}^{(t)},  s^{(t)})=(\mathcal{E}^{(t')},s^{(t')})\Rightarrow (\bm{y}^{(t)}=\bm{y}^{(t')}) $ for all $t'<t$.

Predicting targets $\bm{y}^{(t)}$ requires the model to maintain a \emph{persistent} set of internal states to represent the history of operations for all entities throughout their lifetime.
A na\"{i}ve solution is copying \emph{all} entities after operations, but this quickly becomes intractable due to $\mathcal{O}(tn)$ storage costs---a selective approach is more desirable.
In the following, we present our proposed model, termed \emph{Persistent Message Passing (PMP)}, that combines these desiderata in an efficient way.  See \cref{fig:overview} for an overview.

\paragraph{Overview}
At each step, $t$, our PMP model maintains a set of $N^{(t)}$ \emph{hidden states} $\mathcal{H}^{(t)}=\{\bm{h}_j^{(t)}\in\mathbb{R}^d\}_{j=1}^{N^{(t)}}$.
Initially, $N^{(0)}=n$ in one-to-one correspondence with the $n$ entities, and $\bm{h}_j^{(0)} = \bm{0}$.
During operations, instead of overwriting hidden states, PMP first \emph{selects} a subset of states to persist, and then adds an updated version of these to $\mathcal{H}^{(t)}$---preserving information about previous states in $\mathcal{H}^{(t+1)}$.

To predict outputs, PMP needs to track (i) updated connectivity and (ii) which hidden states are relevant at time-step $s^{(t)}$.
To achieve this, PMP maintains two $N^{(t)}\times N^{(t)}$ adjacency matrices, $\bm{\Pi}^{(t)}$ for \emph{connectivity}  and $\bm{\Lambda}^{(t)}$ for \emph{relevance}.
We initialise $\bm{\Pi}^{(0)}=\bm{\Pi}$ from the task input, and replicate its structure when adding copies of persisted nodes.
By that, we focus on persistency in isolation---even though our framework can easily be extended to inferred $\bm{\Pi}^{(t)}$, e.g. as in \citet{velivckovic2020pointer}.
$\bm{\Lambda}^{(t)}$ is updated so that new nodes link to their old versions.
Note that this form needs $\mathcal{O}(t)$ steps to reach past nodes and we leave improvements to $\mathcal{O}(\log t)$ (e.g.\ \cite{pugh1989skip})  for future work.

PMP closely follows the \emph{encode-process-decode} paradigm  \citep{hamrick2018relational}, with additional \emph{masking} to select  nodes to persist (as done in \citet{yan2020neural,velivckovic2019neural}).

\paragraph{Input encoding}
We encode the current hidden states with both time-stamps and the current operation, using \emph{encoder} networks $f_\text{relevance}$ and $f_\text{operation}$:
\begin{equation}
\label{eq:encoding}
    \bm{v}_j^{(t)} = f_\text{relevance}\left(\mathtt{time\_stamp}(j), t,  \bm{h}_j^{(t - 1)}\right)\qquad
    \bm{z}_j^{(t)} = f_\text{operation}\left(\mathtt{expand}(\mathcal{E}^{(t)}, s^{(t)}), \bm{h}_j^{(t - 1)}\right)
\end{equation}
where, $\mathtt{time\_stamp}(i)$ is the time-step when $\bm{h}_i^{(t-1)}$ was added to the set of hidden states $\mathcal{H}^{(t)}$, and $\mathtt{expand}(\mathcal{E}^{(t)}, s^{(t)})$ maps the $n$ operation features to the $N^{(t-1)}$ hidden states, c.f.\  \cref{sec:experiments}.

\paragraph{Message passing}
The derived representations, ${\bf Z}^{(t)} = (\bm{z}_1^{(t)},  \dots, \bm{z}_{N^{(t)}}^{(t)})$ and ${\bf V}^{(t)} = (\bm{v}_1^{(t)},  \dots, \bm{v}_{N^{(t)}}^{(t)})$ are fed into a \emph{processor network}, $P$, using ${\bm\Pi}^{(t-1)},\bm{\Lambda}^{(t-1)}$ as relational information:
\begin{equation}
    \label{eqn:proc}
    \widehat{\bm{H}}^{(t)} = P\left({\bf Z}^{(t)}, {\bm\Pi}^{(t-1)}\right)\qquad
    \bm{G}^{(t)} = P\left({\bf V}^{(t)}, {\bm\Lambda}^{(t-1)}\right)
\end{equation}
yielding candidate next-step latent features $\widehat{\bm{H}}^{(t)} = (\widehat{\bm{h}}_1^{(t)}, \dots,\widehat{\bm{h}}_{N^{(t)}}^{(t)})$ and relevance latent features $\bm{G}^{(t)} = (\bm{g}_1^{(t)}, \bm{g}_2^{(t)},\dots,\bm{g}_{N^{(t)}}^{(t)})$. $\widehat{\bm{H}}^{(t)}$ are candidates, as $\widehat{\bm{h}}_j^{(t)}$ will be retained only if state $j$ is persisted.

\paragraph{Relevance mechanism}
In order to decide which candidate next-step latent features are used for computing the operation response, we compute a per-state \emph{relevance} mask using the relevance latents and a
\emph{masking} network $\psi_\text{relevance}$ as $\mu_j^{(t)}=\psi_\text{relevance}(\bm{g}_j^{(t)})\in\{0,1\}$.
This mask is used to select relevant hidden states from $\widehat{\bm{H}}^{(t)}$ in downstream computation of response and persistency.

\paragraph{Persistency mechanism}
Many efficient data structures only modify a small (e.g. $\mathcal{O}(\log n)$) subset of the entities at once \citep{cormen2009introduction}.
We explicitly incorporate this inductive bias into PMP: In order to decide which of the candidate next-step  latent features $\widehat{\bm{h}}_i^{(t)}$ are appended to the set of hidden states $\mathcal{H}^{(t)}$, we predict a per-state \emph{persistency} mask (on relevant states only), $\phi_j^{(t)}=\psi_\text{persistency}(\widehat{\bm{h}}_j^{(t)}\cdot \mu_j^{(t)})\in\{0,1\}$.
Using this mask, we generate a set of \emph{next-step latent features} by appending  masked candidates to the previous set of hidden features as
$\mathcal{H}^{(t)} = \mathtt{concat}(\mathcal{H}^{(t-1)}, \{\widehat{\bm{h}}_{j}^{(t)}\}_{j:\phi_j^{(t)}=1}).
$

\paragraph{Updating adjacency}
We subsequently generate next-step adjacency matrices.
For $\bm{\Pi}^{(t)}$, we append copies of masked rows $\{j: \phi_i^{(t)}=1\}$ to the end, and additionally update connections to others replaced nodes within those rows.
For $\bm{\Lambda}^{(t)}$, we simply let each updated node point to its persisted predecessor.
These updated hidden states and adjacency matrices are used in the next time-step.

\paragraph{Readout}
Before moving to the next time-step, we compute a \emph{response} to the input operation from the $\widehat{\bm{H}}^{(t)}$ using a \emph{decoder} network $g$ on \emph{relevant} nodes ($\mu_j^{(t)}=1$),
$\bm{y}^{(t)} = g\left(\bigoplus_{j: \mu_i^{(t)}=1}\bm{z}_j^{(t)}, \bigoplus_{j:\mu_i^{(t)}=1}\widehat{\bm{h}}_j^{(t)}\right),$
where $\bigoplus$ is a \emph{readout} aggregator (we use maximisation).

\paragraph{PMP components}
In our implementation, encoder, decoder, masking and query networks are all linear transformations.
Echoing the results of prior work on algorithmic modelling with GNNs \citep{velivckovic2019neural}, we recovered strongest performance when using \emph{message passing neural networks} \citep{gilmer2017neural} for $P$ in \cref{eqn:proc}: nodes aggregate messages from neighbouring nodes and combine those via an aggregation mechanism (we use maximisation), see \cref{app:training_details} for details.

\paragraph{Optimisation}
Besides the operation response loss for $\bm{y}^{(t)}$, PMP optimises the cross-entropy of relevance and persistency masks $\mu_j^{(t)}, \phi_j^{(t)}$ against the ground-truth.
It is also possible to provide additional supervision for e.g.\ predicting per-node responses from the $\widehat{\bm{H}}^{(t)}$, as done in the experiments.

\begin{figure}
    
    \centering
    \vspace{-1em}
    \includegraphics[width=\linewidth]{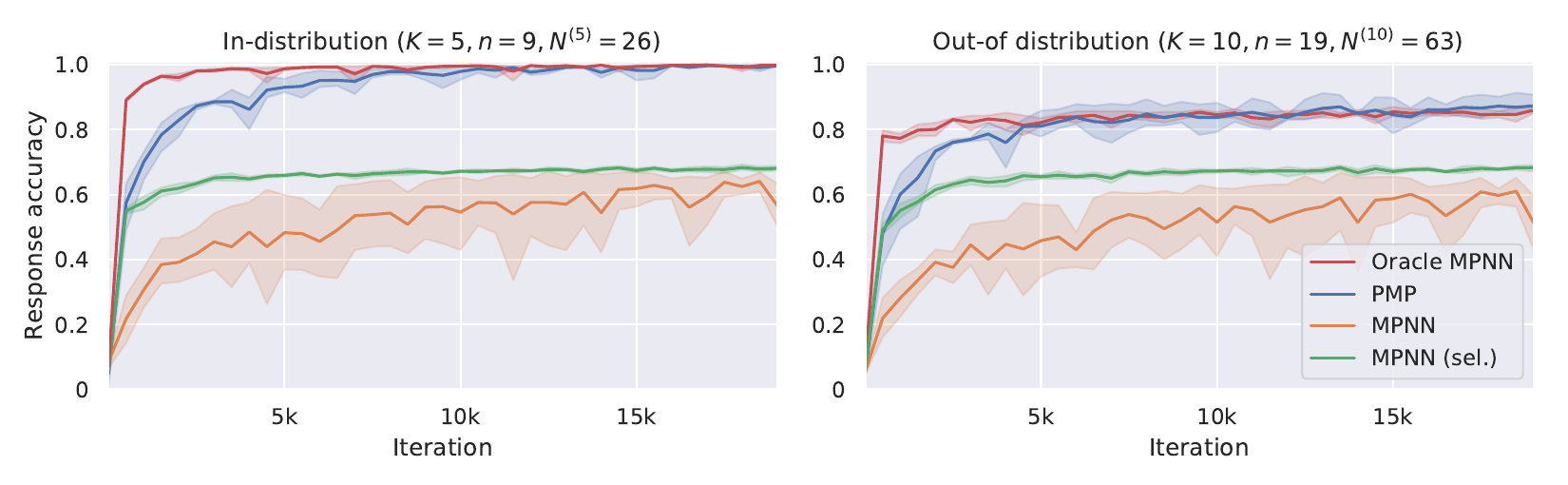}
    \caption{Query response accuracies (all bits correct) for the persistent RMQ task on in-distribution test data ({\bf left}) and on $2\times$ larger test inputs ({\bf right}). PMP gradually catches up to the Oracle MPNN, significantly outperforming the two overwriting baseline MPNNs. Errors bars are over 5 seeds.}\vspace{-1em}
    \label{fig:eval_curves}
\end{figure}

\section{Task: Range minimum query with persistent segment trees}\label{sec:experiments}

We focus on the highly versatile \emph{range minimum query} (RMQ) task on an array $A$ of $K$ integers. At each step $t$, we perform one of two operations: (a) \emph{setting} an individual array element to a new value, $A_k^{(t+1)}\leftarrow x$, or (b) \emph{querying} for the \emph{minimum} value over a particular contiguous range, at a previous point in time, $\min_{a\leq k\leq b}A_k^{(t')}$, for $t'\leq t$. 

RMQ-style tasks are effectively solved with \emph{segment trees} \citep{bentley1977solutions}. Segment trees are almost-complete, full binary trees consisting of $n=2K-1$ nodes, such that leaf nodes correspond to array elements, and any intermediate node maintains the minimum of its two children.
Updating a leaf nodes' value only requires updating the $\mathcal{O}(\log n)$ nodes along the path to the root.
Querying requires minimising over the \emph{canonical} covering of the query range, which contains $\mathcal{O}(\log n)$ nodes. 
Segment trees can be made \emph{persistent} (PST) to for querying of past states: upon update, we copy instead of modifying nodes, resulting in additional $\mathcal{O}(\log n)$ storage per update.

We choose to use PMPs to imitate computations of persistent segment trees in our experiments, not only because of their high importance across computational geometry and combinatorial heuristics, but also because their ``backbone'' tree connectivity remains identical across steps. This allows us to focus on  the utility of PMP's persistency mechanism (with hard-coded connectivity updates).

To model segment trees with PMP, we let our $n$ entities correspond to the PST nodes.
Specifically, PMP is provided with operation representations for (i) updating a leaf node (indicator feature) to a new integer value (global feature), or (ii) querying the minimum value in a given range (indicator feature), at a past state.
Our PMP model is supervised in a \emph{teacher-forced} setup: it is guided to either: (a) select appropriate nodes to persist, and make their latents predictive of updated minimum values, in case of updates; or (b) select the relevant canonical cover of the relevant snapshot of the PST, and produce the appropriate answers over them, in the case of queries. 
We represent all integers in binary form (c.f.\ \citet{yan2020neural}) and consider a query answer correct if \emph{all} bits are correctly predicted. 

\vspace{-.1cm}
\paragraph{Dataset \& training}
We generate a dataset consisting of 10,000 PST rollouts over arrays of size $K=5$, resulting in $N^{(0)}=n=9$ initial nodes.
For each rollout, we perform 5 updates with random position and new value, resulting in $N^{(5)}\approx26$ nodes on average, followed by 5 random queries.
We compute features representations of queries (indicators) and updates (indicators \& binary encodings) dynamically during rollouts, including their mapping onto the model's growing collection of hidden states via $\mathtt{expand}$ in \cref{eq:encoding}.
For in-distribution evaluation, we generate another 200 rollouts from the same distribution. For out-of-distribution evaluation, we in addition use arrays of size $K=10$, and 10 updates, resulting in $n=19$ and $N^{(10)}\approx63$ nodes.
Further details can be found in \cref{app:training_details}.

\vspace{-.1cm}
\paragraph{Results}
In \cref{fig:eval_curves}, we provide the test query accuracy (fraction of all bits correctly predicted) of our PMP model for answering RMQ, in-distribution and over a dataset of more than \emph{twice} as many nodes and operations, \textit{i.e.,} out-of-distribution. At test time, teacher forcing is disabled, and a model must be properly rolled out over the entire \emph{lifetime} of the data structure, including both persistency and relevance mechanisms. 
We find that PMP is close to perfect in predicting masks for both mechanisms (results not shown)---this is crucial as persisting wrong nodes, or selecting wrong relevant nodes would severely corrupt the model's memory.
We compare PMP against message passing networks  (MPNNs) \citep{gilmer2017neural} which \emph{overwrite} rather than persist (either entirely, or selectively by using a node-level mask), and an \emph{oracle} MPNN, which predicts from the \emph{given correct} snapshot of the segment tree and hence doesn't need to memorise any evolution of the nodes.

Our test results are highly indicative of the strong performance of PMP: not only are they consistently more accurate at answering temporal range queries than their overwriting counterparts, but, given enough training time, they are capable of \emph{matching} the oracle GNN even over long rollouts. This demonstrates the robustness of the proposed neural persistency mechanism, and for temporal querying, they appear to effectively relieve the pressure on a GNN's internal representations. 

Interesting directions for future work are improving scalability of the number of updates, dealing with dynamic connectivity structure, and unsupervised learning of relevance and persistency masks.

\bibliographystyle{iclr2021_workshop}
\bibliography{library}

\newpage
\appendix
\appendixpage
\section{Training details}\label{app:training_details}
We generate a dataset consisting of 10,000 PST rollouts over arrays of size $K=5$. The array values are randomly initialized between $[1, 15]$ (represented using a 4 bit binary encoding), by first sampling lower bound from the initial interval, and then sampling array elements uniformly between that and the chosen upper bound.
We choose this distribution in order to ensure non-trivial query responses for minimum queries.

We perform 5 PST updates where we uniformly sample update index $k$ and new value $x$ from the same distribution as the existing values.
This leads to $12,16,19,23,26$ nodes on average, respectively after each update.
After the updates, we perform 5 PST queries, with uniformly sampled query bounds $[a,b]$ and a random past time-step $t'<t$.
We use a shared representation for both queries and updates, setting query features to $\bm{0}$ when computing updates and vice versa.
In the update features, we add an indicator for the node corresponding to the updated PST leaf node, indicators for all leaf nodes, and a global binary encoding of new array value $x$.
For query features, we add an indicator for the  leaf nodes which correspond to the query interval bounds $[a,b]$, and indicators whether nodes are left or right children or root nodes.
Using a 4-bit binary encoding of array values, this leads to a $10$-dimensional representation of operations.

It is straight-forward to dynamically compute these features (of $n$ entities) for the growing set of $N^{(t)}$ hidden states via  $\mathtt{expand}$ in \cref{eq:encoding}: tagging leaf nodes, node type (left/right/root), etc can be inferred from the maintained connectivity of added hidden states within the PMP model.

For in-distribution evaluation, we generate another 200 rollouts from the same distribution. For out-of-distribution evaluation, we in addition use arrays of size 10, and 10 updates resulting in $n=19$ and on average $23,28,32,37,41,45,50,54,58, 63$ nodes respectively after each update. This is followed by 5 queries.

We train PMP for $2\cdot 10^{4}$ iterations using a batch size of $16$ (rollouts) using Adam \citep{kingma2014adam} with a learning rate of $10^{-3}$, re-running the model with 5 different seeds.
Hidden states have $d=64$ dimensions.
We use 10 message passing steps in the processor network, which we found to lead to slightly better performance of PMP compared to fewer steps.
We hypothesise this is due to the linked list approach in implemented in the relevance connectivity matrix $\bf{\Lambda}^{(t)}$---using a data structure that allows search in logarithmic time (such as skip lists, \citep{pugh1989skip}) are likely to allow lower the number of message passing steps.

\paragraph{Message passing network}
PMP builds on message passing neural networks \citep{gilmer2017neural}, used as processor network $P$ in \cref{eqn:proc}. The explicit computation of candidate next-step latent features $\widehat{\bm{h}}_j^{(t)}$ and relevance latent features ${\bm{g}}_j^{(t)}$ is:
\begin{equation*}\label{eqn:mpnn}
    \widehat{\bm{h}}_j^{(t)} = U\left(\bm{z}_j^{(t)}, \max_{{\bm\Pi}^{(t-1)}_{jj'} = 1} M\left(\bm{z}_{j'}^{(t)}, \bm{z}_j^{(t)}\right)\right)\qquad {\bm{g}}_j^{(t)} = U\left(\bm{v}_j^{(t)}, \max_{{\bm\Lambda}^{(t-1)}_{jj'} = 1} M\left(\bm{v}_{j'}^{(t)}, \bm{v}_j^{(t)}\right)\right)
\end{equation*}
where $M$ and $U$ are linear layers producing \emph{vector messages}, followed by ReLU. 

\end{document}